\pdfoutput=1

\documentclass[11pt]{article}

\usepackage{EMNLP2022}

\usepackage[T1]{fontenc}
\usepackage[utf8]{inputenc}
\usepackage{graphicx}
\usepackage{times}
\usepackage{latexsym}
\usepackage{amsmath,amssymb}
\usepackage{booktabs}
\usepackage{multirow}
\usepackage{cellspace}
\usepackage{float}
\usepackage{xspace}
\usepackage{mathtools, cuted}
\usepackage{cleveref}

\newcommand{\todo}[1]{\textcolor{red}{#1}}
\newcommand{\why}[1]{\textcolor{blue}{#1}}

\newcommand{\Exp}{\mathbb{E}}
\newcommand{\x}{{\boldsymbol x}}
\renewcommand{\Pr}{\mathbb{P}}

\title{
Event Causality Identification with Synthetic Control}

\author{Haoyu Wang$^1$\thanks{Work done during internship at Allen Institute for AI.}, Fengze Liu$^1$, Jiayao Zhang$^1$, Dan Roth$^1$, Kyle Richardson$^2$\\
$^1$Department of Computer and Information Science, UPenn\\
$^2$Allen Institute for AI\\
\texttt{\{why16gzl,lclarice,zjiayao,danroth\}@seas.upenn.edu,} \\
\texttt{\{kyler\}@allenai.org}}

\begin{document}
\maketitle
\begin{abstract}
Event causality identification (ECI), a process that extracts causal relations between events from text, is crucial for distinguishing causation from correlation. 
Traditional approaches to ECI have primarily utilized linguistic patterns and multi-hop relational inference, risking false causality identification due to informal usage of causality and specious graphical inference.
In this paper, we adopt the Rubin Causal Model to identify event causality:
given two temporally ordered events, we see the first event as the treatment and the second one as the observed outcome.
Determining their causality involves manipulating the treatment and estimating the resultant change in the likelihood of the outcome.
Given that it is only possible to implement manipulation conceptually in the text domain, 
as a work-around, we try to find a `twin' for the protagonist from existing corpora.
This `twin' should have identical life experiences with the protagonist before the treatment but undergoes an intervention of treatment.
However, the practical difficulty of locating such a match limits its feasibility. 
Addressing this issue, we use the \textbf{synthetic control method} to generate such a `twin' from relevant historical data, leveraging text embedding synthesis and inversion techniques.
This approach allows us to identify causal relations more robustly than previous methods, including GPT-4, 
which is demonstrated on a causality benchmark, COPES-hard.
\end{abstract}

\section{Introduction}
Previous endeavours in event causality identification in text have, to a large extent, depended on feature-based approaches where linguistic patterns serve as a crucial role \cite{beamer2009using, do-etal-2011-minimally, hidey-mckeown-2016-identifying, lai-etal-2022-meci}.
These patterns can roughly indicate causal relations in that causal language is widely used in an informal way in everyday life \cite{imbens2015causal}. 
Without proper manipulation of the potential cause, and comparison between the observed outcome and the intervened outcome, these approaches often identify specious causal relations. 
For example, \textit{``because''} is often considered as a causal indicator \cite{hidey-mckeown-2016-identifying}, yet it might not be rigorous as in the case of ``She got a nice job because she graduated from one of the top universities.'' 
It is very possible that the employer paid more attention to the candidate's ability, rather than just the educational background in offering a job.
Regardless of how these linguistic features are obtained - whether extracted from causal keywords and semantic indications, or obtained from language models - any feature-based approach may be predisposed to bias due to their unreliable causality foundations.
Thus, highly sophisticated methods that rely on multi-hop reasoning on graphs for ECI \cite{cao-etal-2021-knowledge, chen-etal-2022-ergo, liu2023kept, chen-etal-2023-cheer, Pu_Li_Zhao_Wang_Li_Liao_Zheng_2024}, also risk being fundamentally flawed in their conclusions.

If we want to reliably discover event causality, say whether there exists a causal relation between a pair of temporally ordered events $(e_1, e_2)$, we need to manipulate $e_1$ and see if $e_2$ still happens in a `parallel universe' in which $e_1$ does not happen. 
In other words, we want to find a `twin' for the protagonist $p$ in the events, who has identical life experiences with $p$ (i.e., a sequence of events) up to the point when $e_1$ takes place, but instead undergoes an intervention of $e_1$.
\begin{figure*}
    \centering
    \includegraphics[width=0.8\linewidth]{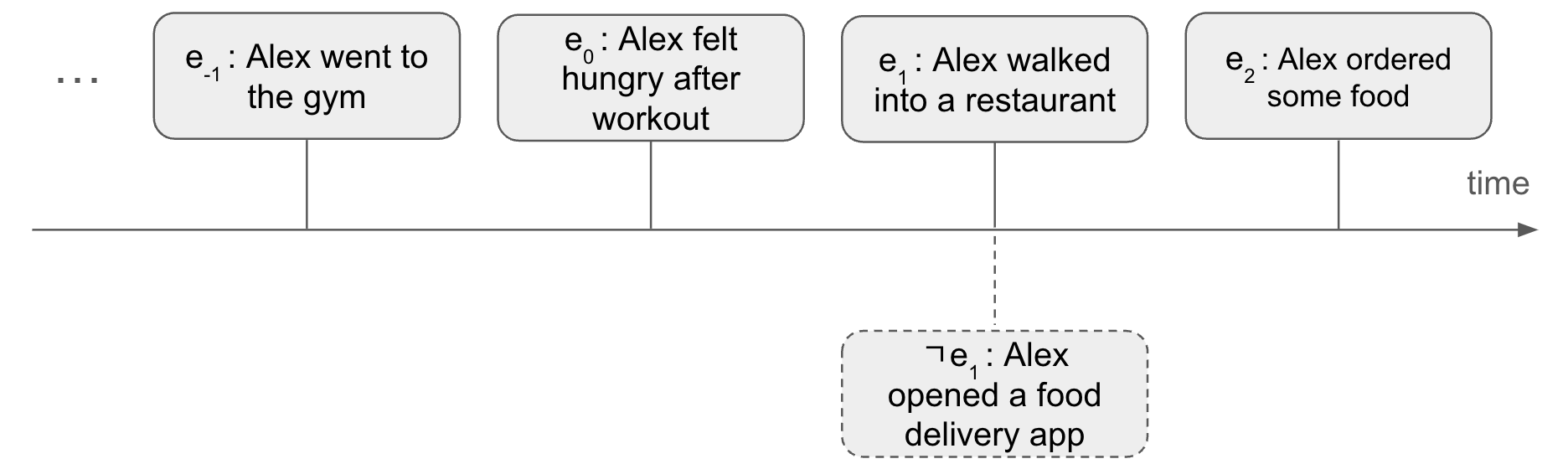}
    \caption{An example illustrating the temporal ordering of treatment event $e_1$, observed outcome $e_2$, and pretreatment events $e_{-1}, e_{0}$ (covariates) on a time axis. To figure out if $e_1$ causes $e_2$, we come up with an intervention of $e_1$, $\neg e_1$, and find that it does not affect the likelihood of $e_2$. And thus, $e_1$ is not the cause of $e_2$.}
    \label{fig:eg}
\end{figure*}
However, it is almost impossible to implement this idea in the text domain (e.g., stories, narratives, and news reports): for any event in text, it is very rare that its protagonist has `twins' that satisfy the aforementioned requirements.
In this work, as a workaround, we attempt to synthesize `twins' by merging relevant event sequences retrieved from a corpus, inspired by a causal inference method, called synthetic control, used in economics and social sciences \cite{abadie2003economic, abadie2010synthetic, abadie2021using}. 
Such event sequence merging is also seen in recent effort in event schema induction in information extraction studies \cite{li-etal-2020-connecting, wen-etal-2021-resin, du-etal-2022-resin, dror-etal-2023-zero}.

Specifically, our approach \footnote{Our code is available at: \url{http://cogcomp.org/page/publication_view/1043}.} consists of three components:
(1) noncontemporary control group retrieval,
(2) control unit synthesis,
and (3) treatment effect estimation.
Given a pair of events with certain context, the first component retrieves relevant event sequences from historical data that can be leveraged to synthesize `twins' via text embedding and inversion techniques \cite{morris-etal-2023-text, morris2024language} in the second component.
And this is followed by the third component which calculates a causal estimand to determine whether there exists a causal relation between the two events.

The proposed methodology fundamentally shifts from conventional ECI methods by introducing the concept of synthetic control to the text domain. 
This allows the inherent linguistic bias which is prevalent in data-driven ECI approaches to be significantly mitigated. 
Moreover, by introducing full-context matching in a continuous space, we overcome the limitation of discrete temporal propensity matching proposed in previous attempts \cite{pmlr-v162-zhang22am, wang-etal-2023-cola} of solving ECI with the potential-outcome framework.
Our approach demonstrates significantly improved results on the COPES-hard dataset, a commonly used causality benchmark, by at least 9\% (relatively)
over existing methods and GPT-4.
The contribution of this paper is threefold: 
\begin{itemize}
    \item Synthetic control method is introduced to solve ECI in text for the first time.
    \item Full-context matching is proposed to synthesize control units with the help of recent language modeling techniques.
    \item Experimental results on the COPES-hard dataset demonstrate the effectiveness and robustness of counterfactual reasoning in text.
\end{itemize}

\section{Preliminaries}
\label{sec:prelim}

In this section, we present the fundamentals of our method, which is grounded on the Rubin Causal Model 
\cite{rubin1974estimating}
and discuss its previous application to the problem of event causality identification in the text domain. And then we discuss the limitation of previous attempts and introduce why we adopt synthetic control in this work.

\subsection{Rubin Causal Model}
The Rubin Causal Model (RCM) is one of the cornerstones of causal inference. 
To illustrate this framework in the text domain, let us consider two temporally ordered events $(e_1, e_2)$ in an article.
They involve a common protagonist, or study unit, $p$, and we want to estimate whether $e_1$ causes $e_2$, with a context that can be modeled as a temporally ordered sequence of event mentions in text: $e_{-t}, e_{-t+1}, \cdots, e_{0}$ (see \Cref{fig:eg} as an example).
Following \citet{pmlr-v162-zhang22am}'s formulation of ECI, we see the first event $e_1$ as the treatment, and second event $e_2$ as the observed outcome. 
To measure the treatment effect, we need to compare the study unit with a control group within which the control units did not undergo event $e_1$, and estimate the change of the likelihood of $e_2$ \textit{had} $e_1$ \textit{been intervened}:
\begin{equation}
\label{eq:task_definition}
\Delta = \mathbb{P} ( e_1 \prec e_2 ) - \mathbb{P} ( \neg e_1 \prec e_2 ).
\end{equation}
Here we use $\prec$ to indicate that $e_1$ occurs before $e_2$, and $\neg e_1$ to denote a manipulation of $e_1$, which can only be conceptual or imaginary.

\subsection{Temporal Propensity Matching}
The most significant challenge in formulating ECI as described above is the spurious correlations introduced by pervasive confounding co-occurrences.
They need to be eliminated for an unbiased estimation of the causal estimand introduced in \Cref{eq:task_definition}.
This can be done by balancing events that precede
$e_1$, or \textit{covariates}. 
Several techniques for balancing covariates \cite{cochran1965planning, rosenbaum1983central, pearl1995causal} have been proposed, e.g., sub-classification, matched sampling, covariance adjustment, and propensity score. 
\citet{pmlr-v162-zhang22am} propose to use \textit{temporal propsensities} for covariate balancing in text.
To this end, \Cref{eq:task_definition} can be rewritten as 
\begin{equation}
    \label{eq:conditional_ate}
    \begin{aligned}
	\Delta =  \Exp_{\x} \left[ \Pr(e_1 \prec e_2 \vert \x)
		- \Pr(\neg e_1 \prec e_2 \vert \x) \right],
    \end{aligned}
\end{equation}
and here the treatment assignment is strongly ignorable
with respect to the covariates $\x$ = $[e_{-t}, e_{-t+1}, \cdots, e_{0}]$.
The propensity score, 
\begin{equation}
    \label{eq:propensity}
    \begin{aligned}
	p(x) = \Pr(e_1| \x),
    \end{aligned}
\end{equation}
is the probability of $e_1$ occurring at time 1 conditioning on the covariates being $\x$ at time equal to or less than 0 (prior to the time $e_1$ happens).
To incorporate the context of $e_1$, \citet{wang-etal-2023-cola} design a mechanism to sample diversified covariates
from multiple timestamps and also use temporal propensity for balancing.
Yet they merge covariates to construct the final covariate set which would lose the temporal interaction within the sequence of context events.

\subsection{Better Context Modeling with Synthetic Control}
Synthetic control is a widely-used method in econometrics and social sciences for policy evaluation and causal inference in observational studies \cite{abadie2003economic, abadie2010synthetic}. It addresses the challenge of having to estimate the counterfactual, a critical aspect in the study of causality. The method involves constructing an artificial control unit – a synthetic control – as a weighted combination of potential control units, rather than relying on just a single control unit. This synthetic control then acts as the counterfactual, representing what would have happened in the absence of the treatment. The causal effect is subsequently estimated by comparing the study unit and the synthetic control unit.  This technique allows for robust treatments of causal effects where finding an event sequence that perfectly mirrors the treated case is impractical. 
\begin{figure}
    \centering
    \includegraphics[width=1\linewidth]{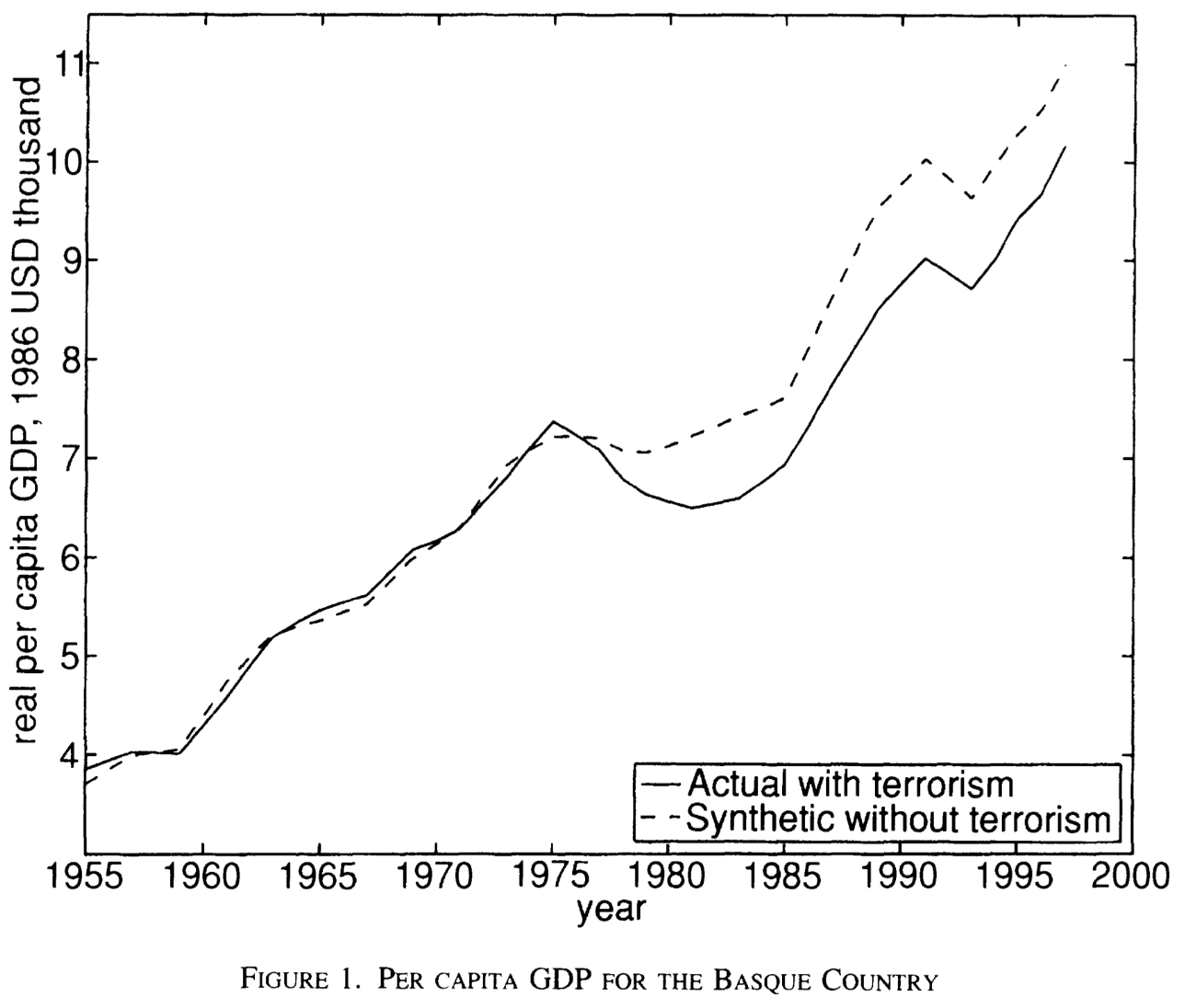}
    \caption{After the outbreak of terrorism in the late 1960's, per capita GDP in the Basque Country declined about 10 percentage points relative to a synthetic control region without terrorism. Figure from \citet{abadie2003economic}.
    }
    \label{fig:basque}
\end{figure}
As illustrated in \Cref{fig:basque}, this method involves creating a synthetic control group from a weighted combination of multiple untreated units that closely mimic the pre-intervention characteristics and trends of the treated unit. In this case, the solid line represents the actual per capita GDP of the Basque Country, which experienced the impact of terrorism in the late 1960's, while the dashed line represents the synthetic per capita GDP constructed from adjacent regions unaffected by terrorism. By comparing the actual GDP with the synthetic GDP after the onset of terrorism, the graph visually depicts the negative economic impact of terrorism on the Basque Country. This gap between the lines highlights the divergence from what the economic trajectory might have been in the absence of terrorism, thereby demonstrating the usefulness of the synthetic control method in assessing causal effects. 

With the synthetic control method, we can model longer context in the RCM framework while maintaining the temporal structure of the original event sequence.
Yet in the text domain, it is harder to find contemporary control group like those GDP curves of adjacent regions.
In the following sections, we further discuss how we perform the synthetic control method in the context of event causality identification in text, by retrieving  noncontemporary control groups and synthesizing control units from them.
\section{Method}
For a pair of events $e_1$ and $e_2$ mentioned in some context that we consider as the study unit, we want to 
(1) find relevant stories (event sequences) from historical data that can be considered as noncontemporary control group, and 
(2) merge them to create a synthetic control unit, and then 
(3) calculate the causal estimand. 

\subsection{Noncontemporary Control Group Retrieval}
Since it is very rare that `twins' of the protagonist exist in some existing corpus, we turn to noncontemporary articles of the same topic that happen not necessarily at the same time as the study unit.
Even though these articles do not form a perfect control group, we can filter and obtain the most relevant ones and merge them as a synthetic control unit (see \Cref{label:cu_synthesis}).

As a preprocessing step, we first use \texttt{gpt-3.5-turbo}\footnote{\url{https://platform.openai.com/docs/models}} to anonymize the entire event sequence so it does not contain any specific entities\footnote{See \Cref{sec:anonymization} for detailed prompt.}.
For example, in the event description, we blur the entities: we convert ``Timmy'' to ``a boy;'' ``Mary'' into ``a girl.''
The reason this operation is 1) our focus is event. 2) we admit that arguments, especially people, play important roles in the progress of an event. 
But it is also the actions that define a person's character. 
Too much information about the arguments might mislead the retrieval process and subsequently the creation of synthetic control. 
However, we do not use abstraction\footnote{This insight comes from our experiments where the performance worsens as the level of abstraction increases, e.g. from best to worst, in terms of performance of \texttt{gpt-3.5-turbo}, `Tom' $>$ `a boy' $>$ `a person'.} when we determine the similarity of sentences using \texttt{gpt-3.5-turbo}.
\begin{figure*}
    \centering
    \includegraphics[width=0.85\linewidth]{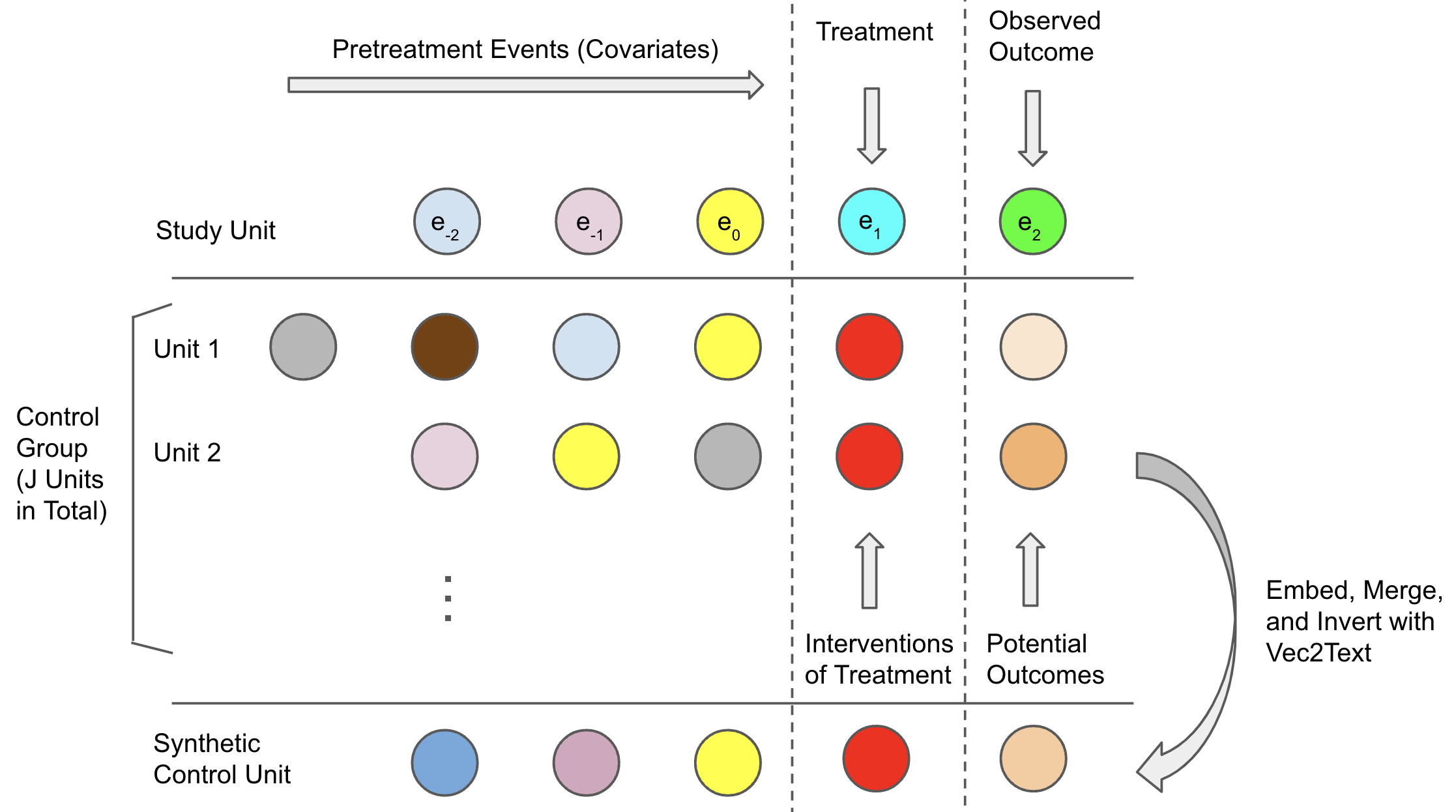}
    \caption{An illustration of how our approach works. The first row shows the event sequence of the study unit, followed by nontemporary control group below (Unit 1, Unit 2, $\cdots$) retrieved from a large corpus. These units are then merged in the embedding space to create a synthetic control unit shown in the last row.
    }
    \label{fig:1}
\end{figure*}
Then we use BM25 \cite{robertson2009probabilistic} to retrieve $n$ relevant documents from a large corpus that has a good amount of topic coverage,
given these event descriptions from the study unit. 
Yet not all of these documents satisfies our requirements: (1) we need the pretreatment events of the study unit and the control group to be as close as possible; (2) the units in the control group cannot contain the treatment event, but intervention of treatment instead.
We do the same preprocessing procedure with the retrieved documents and use \texttt{gpt-3.5-turbo} to summarize\footnote{See \Cref{sec:sum} for detailed prompt.} the retrieved documents \cite{zhang2023benchmarking}. 
These pieces of text are embedded into vectors using \texttt{text-embedding-ada-002}\footnote{\url{https://platform.openai.com/docs/guides/embeddings}}. 
Leveraging the embeddings, we keep those documents with pretreatment events whose cosine distance is higher than a certain threshold. However, measuring event similarity with cosine similarity can be rather arbitrary at times \cite{Steck_2024}. For example, ``A person loves food'' and ``A person does not love food'' can have a cosine similarity of $>0.9$, depending on the specific embedding model used. As such, cosine similarity is only used as a first round of filtering and we subsequently examine the similarity of kept documents\footnote{See \Cref{sec:similar} for detailed prompt.} using \texttt{gpt-3.5-turbo}. 

There are three key parts of event similarity that we check using \texttt{gpt-3.5-turbo}: (1) Pretreatments of the kept documents vs. treatment of the study unit. This is done to ensure that the treatment of the study unit does not take place in the pretreatments, which will affect our assessment of the causal estimand. (2) Interventions of the kept documents vs. treatment of the study unit. Due to the arbitrary nature of the cosine similarity measure, we have to ensure that the interventions and treatment are in fact dissimilar. (3) Outcomes of the kept documents vs. treatment of the study unit. Similar to (1), having an intervention similar to the treatment will make our estimates inaccurate. To do this, we independently prompt \texttt{gpt-3.5-turbo} with two slightly different questions:\\

\noindent {\fontfamily{cmss}\selectfont
Ignoring the specific characters, ["does a similar event to event B take place in event A", "is event B a subset of event A"]?
}\\

In this step, the unanonymized texts for both the study unit and the retrieved documents are provided to \texttt{gpt-3.5-turbo} for comparison, since we observe less robust performance as the level of abstraction increases. In the prompt above, ordering of the two input sentences are crucial. Segments from retrieved documents are labeled as event A, and treatment of the study unit is labeled as event B. Instead of asking if the two events are similar, we intentionally phrase the prompt in such a way to look for any indication of event B within event A. For example, there might be many subevents in the pretreatments of a retrieved document, but we only need to make sure that there is no subevent that is similar to the treatment of the study unit. One drawback of this measure, however, is the hallucination of \texttt{gpt-3.5-turbo} even when temperature has been set to zero. For example, when event A=``A loyal dog named Buddy played in the mud and got very dirty.'' and event B=``Timmy got in the tub and his mom bathed him.'', the two questions above output true with the reasoning being ``They both involve getting clean by taking a bath'', likely due to co-occurrence bias \cite{kang2023impact}. However, this prompting approach still gives the desired result for this step, which is high recall on detecting events similar to the original treatment. Therefore, we obtain the responses $r_1$ and $r_2$ and output the final similarity as $(r_1 \vee r_2)$. 

We require the number of kept documents to be $\ge 2$ in order to proceed to synthesizing the control unit, otherwise, our approach outputs ``indeterminate'' due to the limited size of the corpus. 

\subsection{Merging Control Group}
\label{label:cu_synthesis}
After we find the relevant control group $[U_1, U_2, \cdots, U_{J}]$ as shown in \Cref{fig:1}, we embed the anonymized pretreatments of the study unit and the control units using embedding function 
$\phi$ = \texttt{text-embedding-ada-002} and obtain text embeddings $u_{\text{study}}$ and $[u_1, u_2, \cdots, u_J]$, respectively. If the treatment being tested is the first sentence in the sequence, we prompt \texttt{gpt-3.5-turbo} to generate an augmented context as the ``pretreatment''. 
We then apply ridge regression to find some optimal weights $w_1, w_2, \cdots, w_J$ such that
\begin{equation}
    \label{eq:ridge}
    \begin{aligned}
	(u_{\text{study}} - \Sigma^J_{j=0} w_j \cdot u_{j})^2 + \lambda \Sigma^J_{j=1} w_j^2
    \end{aligned}
\end{equation}
is minimized. The $L_2$ regularization is added to prevent overfitting to any single retrieved document. The same set of weights is then applied to the outcomes of the retained documents to produce a single embedding vector.
\subsection{Control Unit Synthesis}
We use the weights obtained to  linearly combine the embedding vector of the outcome in each control unit $[o_1, o_2, \cdots, o_J]$
\begin{equation}
    \label{eq:combine}
    \begin{aligned}
	o_{\text{synthetic}} = \Sigma^J_{j=0} w_j \cdot o_{j}.
    \end{aligned}
\end{equation}
Then the combination is inverted to generate the synthetic potential outcome in a textual format using a Vec2Text function.
The state-of-the-art Vec2Text function proposed by \citet{morris-etal-2023-text} is built to iteratively reconstruct text from its embeddings by treating the inversion problem as controlled generation. It refines an initial text hypothesis through repeated corrections, using the differences between the target embedding and the hypothesis embedding to guide these updates, achieving high accuracy in recovering the original text from dense embeddings.
With such function $\phi^{-1}$, we obtain the inverted text as the synthetic control unit in its textual format:
\begin{equation}
    \label{eq:invert}
    \begin{aligned}
	\text{inverted text} = \text{LM}^{-1}(o_{\text{synthetic}})
    \end{aligned}
\end{equation}

\subsection{Causal Estimand}
The similarity of the synthesized outcome (Event A) and the original outcome (Event B) are assessed with \texttt{gpt-3.5-turbo} using the same prompt as the filtering process\footnote{See \Cref{sec:similar} for detailed prompt.}.
Since the output of the Vec2Text inversion captures only a vague idea of all the outcomes of the top retrieved documents, our prompt encourages \texttt{gpt-3.5-turbo} to ``fill in the blanks'' and evaluate whether $e_2$ is present in the synthetic outcome. For example, when
\textbf{Event A (synthetic outcome)} is
``The mom and dad drank a cup of coffee. The little mouse was tired, and the mom sat down. They greeted each other, and enjoyed the coffee together. The parents were happy, and the little mouse sat down. The mom sipped a cup of coffee, and the child felt better'';
\textbf{Event B (observed outcome $e_2$)} is ``After i was done, i felt much better'', 
our prompt outputs true since both Events A and B involve someone feeling better.
And this is a scenario where hallucination of Large Language Models (LLMs) \cite{rawte-etal-2023-troubling, mckenna-etal-2023-sources} is helpful in reasoning, since the text recovered from  embedding is sometimes incomprehensible for human beings but comprehensible for LLMs themselves.

\section{Experiments}
We conduct experiments to demonstrate the effectiveness of our proposed approach.
\subsection{Dataset}
For our evaluation of event causality identification in text, we leveraged the Choice of Plausible Event in Sequence (COPES) dataset \cite{wang-etal-2023-cola}, one of the event causality identification benchmarks. The COPES dataset was assembled via Amazon Mechanical Turk and includes event sequences extracted from ROCStories \cite{mostafazadeh-etal-2016-corpus}, where each sequence holds five chronologically ordered events. The annotators were tasked to identify whether a given event was causal to the final event in the sequence. COPES, with its emphasis on causality and chronological event sequencing, serves as an ideal testbed for our focus - integrating the potential outcome framework and synthetic control method into the realm of textual ECI.

Although LLMs have shown relatively strong performance at many causal reasoning tasks, many have argued that LLMs are just ``causal parrots'' \cite{zecevic2023causal} and lack a genuine comprehension of the causal framework \cite{ashwani2024cause}. 
Therefore, our focus is on a subset of the COPES data whose causal relationships are difficult for LLMs to grasp in a zero-shot setting.
Specifically, out of the 340 samples from the COPES dataset, there are 70 samples in total which show $\ge 3$ false positives when \texttt{gpt-4-turbo} is prompted to identify the possible cause(s) in a zero-shot setting. One of the 70 samples is shown below: 

\noindent \textbf{Events:} `Denise loved playing pokemon go.', `She decided to take a walk so she could play.', `While she was crossing the street, denise saw a pokemon on her screen.', `Denise was almost hit by a car as she walked into traffic.', `She decided to only play on the sidewalk from now on.' \\
\textbf{Outcome:} `She decided to only play on the sidewalk from now on.' \\ 
\textbf{Cause:} `Denise was almost hit by a car as she walked into traffic.' 

In the example shown above and under a zero-shot setting, \texttt{gpt-4-turbo} identifies all four event sequences that preceed the observed outcome to be causes. While all four sequences might co-occur frequently with the observed outcome, narrowing down to the one true cause requires a more robust framework. The goal of our approach is to improve the precision without too much deterioration in recall, thereby achieving an increase in the F1-score. 

\subsection{Baseline Methods}

\begin{itemize}
    \item \textbf{Direct prompting}: given the five chronologically ordered events, we ask \texttt{gpt-4-turbo} to select event(s) from the first four that cause(s) the fifth event.
    \item \textbf{Prompting with counterfactual thinking}: One by one, we ask \texttt{gpt-4-turbo} if the fifth event would still happen, had each of the first four events not happened\footnote{See \Cref{sec:counter} for detailed prompt.}.
    \item \textbf{ROCK}: A RCM based causal inference framework \cite{pmlr-v162-zhang22am} that generates interventions and balances covariates with temporal propensity matching.
    \item \textbf{COLA}: A RCM based causal inference framework \cite{wang-etal-2023-cola} that generates interventions and balances covariates from multiple timestamps so as to take context information into account.

\end{itemize}

\subsection{Experimental Setup}

Since the COPES dataset consists of primarily children's stories, we use TinyStories \cite{eldan2023tinystories} which resembles the content of the samples as our corpus. 
The choice of TinyStories as the corpus for retrieval is mostly as a result of the nature of our test dataset, but the approach of synthesizing control units from a large corpus also applies to identifying causal relationships from real life events based on retrieval from narratives and news corpus, among other genres. 

During experimentation, we set the corpus retrieval size $n$ to be 100. The maximum number of documents kept for inversion is 5, and the minimum is 2, i.e. if we are unable to find at least 2 documents that satisfy our criteria, the algorithm outputs ``indeterminate'' for the event pair. The cosine similarity threshold is set to 0.8 for both pretreatment similarity and treatment dissimilarity. For ridge regression, we set the parameter $\lambda$ to 1.0. When we apply Vec2Text to generate the synthetic potential outcome, we set the number of steps to 10 with a beam width of 4. 


\subsection{Results}
Table \ref{tab:performance_summary} below summarizes the performance of our algorithm compared against two previous RCM based methods     and zero-shot performance of \texttt{gpt-4-turbo} and \texttt{gpt-4-turbo} with counterfactual thinking.

\begin{table} [hbt!]
    \centering
    \begin{tabular}{cccc}
    \hline
    \hline
         & Precision & Recall & F1 \\
    \hline
        \texttt{gpt-4-turbo} & 0.2052 & 0.8462 & 0.3303 \\
        Counterfactual & 0.1566 & 0.9013 & 0.2668\\
        ROCK & 0.2239 & 0.6960 & 0.3388 \\
        COLA & 0.2437 & 0.8643 & 0.3802 \\ \hline
        Synthetic Control & 0.2663 & 0.75 & 0.3930 \\
    \hline \hline
    \end{tabular}
    \caption{Comparison of model performances on the COPES-hard dataset. 
    }
    \label{tab:performance_summary}
\end{table}



Our Synthetic Control approach delivers a remarkable precision of 0.2663, marking a significant rise of 29.8\%, or roughly six percentage points, over the precision achieved by direct prompting \texttt{gpt-4-turbo}. It also shows a remarkable improvement over other models such as ROCK (0.2239) and COLA (0.2437), reinforcing the accuracy of our method in distinguishing true causal relationships and reducing false positives.
Moreover, this approach reflects a 19.0\% enhancement in the F1-score compared to \texttt{gpt-4-turbo}, thus highlighting a more balanced performance between precision and recall. 
Notably, our results indicate that less compute and parameter-intensive models, such as \texttt{gpt-3.5-turbo}, can outmatch larger models in discerning causal relationships within text. This underscores that the efficiency of a model in handling causality-related tasks is not strictly dependent on its size or complexity.

In conclusion, our synthetic control approach provides a robust method for event causality identification in the text, underscoring broad-ranging improvements across standard performance metrics relative to existing approaches, and demonstrating the potential superiority of leaner models.



\section{Related Work}
\subsection{Causal Inference}

Causal inference has been a pivotal area of study in both statistics and artificial intelligence. Two dominant frameworks have emerged in this field: the Rubin Causal Model (RCM) and Pearl's do-calculus.
The Rubin Causal Model, also known as the potential outcomes framework, was developed by \citet{neyman1923}, \citet{rubin1974estimating}, and \citet{holland1986statistics} and is grounded in the idea of counterfactuals. In this model, causality is established by comparing potential outcomes—what would happen both with and without the treatment. This approach relies heavily on randomized controlled trials (RCTs) to estimate causal effects, providing a clear mechanism to distinguish causation from correlation. Key methodologies within this framework include propensity score matching \cite{rosenbaum1983central, ho2007matching} and synthetic control methods \cite{abadie2010synthetic, billmeier2013assessing, saunders2015synthetic}, which are particularly useful in observational studies where randomization is not feasible.

Different from the potential outcome framework, \citet{pearl1995causal}'s do-calculus is rooted in structural causal models (SCMs) and utilizes directed acyclic graphs (DAGs) to represent causal relationships. The do-calculus provides a formal language to express and manipulate these relationships, offering tools to calculate causal effects from observational data by simulating interventions \cite{pearl2009causality}. This framework has been instrumental in formalizing causal inference, especially in scenarios where RCTs are not possible, and has broad applications across various domains, including epidemiology, social sciences, and artificial intelligence.
\subsection{ECI in NLP}
Event causality identification in natural language processing (NLP) has traditionally relied on feature-based approaches, where linguistic patterns are key indicators of causal relations. Early works focused on extracting causal relationships using predefined causal markers such as ``because,'' ``therefore,'' and ``due to'' \cite{beamer2009using, hidey-mckeown-2016-identifying}. However, these approaches often fall short in distinguishing causation from correlation, as causal language in everyday text can be used informally and ambiguously \cite{imbens2015causal}.
Recent advancements have shifted towards leveraging deep learning and graph-based methods to improve ECI. Multi-hop reasoning on graphs and the integration of external knowledge bases have shown promise in enhancing the accuracy of causality extraction \cite{cao-etal-2021-knowledge, chen-etal-2022-ergo}. Despite these improvements, these methods still face challenges related to bias and the reliability of inferred causal relations, particularly when relying heavily on linguistic patterns without robust causal foundations.

Two recent work, ROCK \cite{pmlr-v162-zhang22am} and COLA \cite{wang-etal-2023-cola}, mitigate the aforementioned bias by applying  the potential outcome framework to ECI.
ROCK introduces temporal propensity matching to construct intervention of treatments, whereas COLA improves upon ROCK by considering the context of events at the same time. 
Yet COLA is still limited by its coarse modeling of context events, i.e., ultimately merging covariates to construct a covariate set, which would lose the temporal interaction and sequential information within the context events.
Moreover, both methods adopt intervention generation with language models which is somewhat problematic given the prevalent hallucination issue \cite{mckenna-etal-2023-sources, rawte-etal-2023-troubling} in LLM generation.
In contrast, our approach not only models the context with text embedding in the continuous space, but also retrieves from reliable sources instead of relying on LLM generation.


\subsection{Embedding to Text}
The process of recovering text from language model (LM) embeddings \cite{adolphs-etal-2022-decoding, ram-etal-2023-token}, also known as LM inversion, has gained significant attention with the rise of deep learning and transformer-based models in NLP. Text embeddings, such as those produced by BERT \cite{devlin-etal-2019-bert}, GPT-2 \cite{radford2019language}, and other transformer models, encapsulate semantic information in dense vector representations. These embeddings are instrumental in a variety of NLP tasks, including text classification, machine translation, and question answering.
However, the challenge of reversing these embeddings back into human-readable text, or LM inversion, is crucial for interpretability and for applications like counterfactual generation in causality studies. Recent research has explored various techniques for this inversion process. For instance, \citet{morris-etal-2023-text, morris2024language} leverage neural networks to decode or generate text from its embeddings, ensuring that the generated text closely matches the original semantic meaning.
In text-based causal inference tasks, embeddings can be used to generate synthetic control units by constructing `twins' for protagonists of events. By synthesizing events and contexts that are statistically similar to those experienced by a protagonist, we can estimate causal effects in scenarios where direct manipulation is impractical.

\section{Conclusion}
Our work shows that creating counterfactuals with synthetic control, a concept that has been widely adopted in other disciplines such as economics, can be effectively applied to event causality identification under zero-shot settings. This retrieval-based method instills more confidence in the result, offering more robust performance in tasks at which state-of-the-art LLMs might fail. Our results also open up opporunities for research with more complex datasets and causal relationships.

\section*{Limitations}
Our research has made significant advancements in event causality identification in text using the synthetic control method. However, it is essential to acknowledge the limitations.

The first significant limitation of our approach hinges on the quality and relevance of the retrieved control units. The synthetic control method's accuracy highly depends on the available pool of control units drawn from historical data. If the data lacks adequate and suitable counterparts for the treatment group or is biased towards certain types of sequences or events, it may hamper the function and outcomes of the model.

The time complexity of our method could be another limitation. The process of retrieving relevant control units, synthesizing synthetic controls, and estimating causal effects can be computationally intensive and time-consuming, especially when dealing with large datasets. The scalability of the method is a factor that needs further considerations to make it feasible for larger-scale applications. 
Our method also relies heavily on text embeddings for the synthesis of control units. Despite their proficiency at capturing semantic information from text, the embeddings generated by language models are not perfect and could inadvertently introduce a level of semantic loss or distortion. The process of recovering the text from the embeddings, also mentioned as model inversion, is also prone to error and could affect the quality of the generated 'twins'. 
Our approach currently assumes that the event sequences are independent and identically distributed, which might not hold in many real-world scenarios. For instance, in a narrative, events usually have dependencies, and ignoring relationships between sequences can lead to misleading conclusions.


While these limitations present challenges, they also provide directions for future work to enhance our understanding of the application of synthetic control method in identifying event causality in text and scale this approach for broader usage within the field.

\section*{Ethics Statement}
Our work involves leveraging machine learning algorithms to enhance the identification of causal relationships in textual data, specifically focusing on event causality. Our primary source of data is the publicly available COPES dataset, which does not involve data of a personal or sensitive nature. 

While the development and application of our approach do not involve immediate ethical concerns, there could arise potential implications in its broader applications. Event causality identification in text could be used in various scenarios, such as content generation, recommender systems, and even legal contexts. It is important to outline possible misuse. 
Firstly, the algorithm can become a tool for spreading misinformation or generating biased content if the causal inferences it draws from the input text are incorrect or misleading. Stringent validation methods and unbiased, accurate control units are essential to mitigate such concerns. 
Secondly, it is critical to be aware of potential biases in the historical data used for retrieving control units. This could impact the development of synthetic controls and subsequently skew the interpretation of causality. 
Lastly, privacy concerns could arise if the method is applied to text that holds private or sensitive information. As researchers, we ought to uphold the privacy and anonymity of any subjects used in such data. 


\section*{Broader Impact}
In this work, we propose a novel approach to event causality identification in text, combining the potential outcome framework and synthetic control method. This research contributes noteworthy advancements in Natural Language Processing and has the potential for substantial broader impacts in various domains.

Our method provides a scientifically rigorous approach to understanding causality in narratives. It opens avenues for greater exploration and understanding in the domain of causal inference from text, which can be critical for fields like social sciences, psychology, law, and many more.
The application of our method could also greatly enhance the development of AI and machine learning models that require proficiency in understanding, figuring out and interpreting event causality. This includes recommendation systems, chatbots, virtual assistants, and AI narrative generation.
Moreover, our synthetic control approach can significantly benefit information retrieval systems, text summarization, text simplification, and information extraction applications. Better understanding of textual event causality could enhance the relevance and quality of queried information.

While there are significant benefits, some potential negative impacts also warrant attention. Causality identification in text can be used to infer sensitive information in adversarial settings, which can pose privacy concerns. Furthermore, the algorithm can unintentionally propagate or intensify existing bias in the data, leading to ethical and social implications in decision-making systems.
The true broader impact of our research will heavily depend on the contexts and domains within which it is applied. Adopting a responsible, ethical, and fair use perspective is vital to maximize the potential benefits while minimizing harm. We encourage future applications to consider these aspects while exploiting this method.

\section*{Acknowledgements}
This research is based upon work supported in part by ONR Contract N00014-23-1-2417, and by the Oﬃce of the Director of National Intelligence (ODNI), Intelligence Advanced Research Projects Activity (IARPA), via IARPA Contract No. 2019-19051600006 under the BETTER Program. The views and conclusions contained herein are those of the authors and should not be interpreted as necessarily representing the oﬃcial policies, either expressed or implied, of ODNI, IARPA, the Department of Defense, or the U.S. Government. The U.S. Government is authorized to reproduce and distribute reprints for governmental purposes notwithstanding any copyright annotation therein.

\bibliography{anthology,custom}

\appendix

\section{Example Appendix}
\label{sec:appendix}
\subsection{Prompt for summarization}
\label{sec:sum}
You will be given a short story. Please help to summarize the key events in the text to 5 or fewer sentences of less than 15 words each.
The events should be in chronological order, and the events should capture the key actors, location, causes, and effects of the event being described.
Return your answer in JSON as a array of strings in the key `result`.

Here is an example:
Text:
```
Once upon a time, there was an ugly frog. The ugly frog lived in a small pond. The frog liked to get things. He would get things from the bottom of the pond. One day, he saw a shiny weight.
The ugly frog wanted the shiny weight. He tried to get it, but it was too heavy. He tried and tried, but he could not get it. The ugly frog was sad. He wanted the shiny weight so much.
Then, a big fish came. The big fish saw the ugly frog and the shiny weight. The big fish wanted to help. The big fish and the ugly frog worked together to get the shiny weight. They were happy to have the shiny weight. They became good friends.
```

Answer:
```
{{
  "result": [
    "An ugly frog who liked to get things lived in a small pond.",
    "One day, the ugly frog saw a shiny weight, and wanted to get it, but could not.",
    "A big fish came, and the fish wanted to help the ugly frog get the shiny weight.",
    "The big fish worked together with the ugly frog to get the shiny weight.",
    "The fish and the frog were happy to get the weight and became good friends."
  ]
}}
```

Now your turn:
Text:
"{text}"

\subsection{Prompt for measuring similarity}
\label{sec:similar}
Given two separate events:

------
Event A: "{event}"
------
Event B: "{test\_event}"
------

Ignoring the specific characters, {question}? Provide your answer in JSON with the keys `is\_similar` and `reasoning`.

\subsection{Prompt for anonymization}
\label{sec:anonymization}
You will be given a story. Your job is to anonymize the names of persons, and replace them with a generic term. If there is nothing to anonymize, return the story as is.

For example, "Mary" should be replaced by "a girl", and "Tim" should be replaced by "a boy". 

Return your result as a string in the key `result` of a JSON object.

Now your turn:
Story: {event}

\subsection{Prompt for counterfactual thinking}
\label{sec:counter}
Here is a story with five events: \{story\}.

Your task is to tell if the \{i\}-th event \{event 1\} is the cause of the fifth event \{event 2\}.

Please think step-by-step. You need to imagine a scenario where the \{i\}-th event \{event 1\}  is intervened by some other event and then determine if the fifth event \{event 2\} would still happen. If the fifth event would still happen, then answer no; else answer yes.

Now tell me if there exists a causal relation between the two events.

\end{document}